\documentclass[lettersize,journal]{IEEEtran}
\usepackage{amsmath,amssymb,amsfonts}
\usepackage{textcomp}
\usepackage{graphicx}
\usepackage{cite}
\usepackage{hyperref}
\usepackage{booktabs}
\usepackage{multirow}
\usepackage{makecell}
\usepackage[table]{xcolor}
\definecolor{lightgray}{gray}{0.9}

\begin{document}

\title{MedDiff-FM: A Diffusion-based Foundation Model for Versatile Medical Image Applications}

\author{Yongrui Yu, Yannian Gu, Shaoting Zhang, and Xiaofan Zhang
\thanks{Corresponding authors: Shaoting Zhang; Xiaofan Zhang.}
\thanks{Yongrui Yu and Yannian Gu are with the Qing Yuan Research Institute, Shanghai Jiao Tong University, Shanghai, China.}
\thanks{Shaoting Zhang is with the Qing Yuan Research Institute, Shanghai Jiao Tong University, Shanghai, China, and also with the SenseTime Research, Shanghai, China (e-mail: zhangshaoting@sensetime.com).}
\thanks{Xiaofan Zhang is with the Qing Yuan Research Institute, Shanghai Jiao Tong University, Shanghai, China, and also with the Shanghai Innovation Institute, Shanghai, China (e-mail: xiaofan.zhang@sjtu.edu.cn).}}

\maketitle

\begin{abstract}
Diffusion models have achieved significant success in both natural image and medical image domains, encompassing a wide range of applications. Previous investigations in medical images have often been constrained to specific anatomical regions, particular applications, and limited datasets, resulting in isolated diffusion models. This paper introduces a diffusion-based foundation model to address a diverse range of medical image tasks, namely MedDiff-FM. MedDiff-FM leverages 3D CT images from multiple publicly available datasets, covering anatomical regions from head to abdomen, to pre-train a diffusion foundation model, and explores the capabilities of the diffusion foundation model across a variety of application scenarios. The diffusion foundation model handles multi-level integrated image processing both at the image-level and patch-level, utilizes position embedding to establish multi-level spatial relationships, and leverages region classes and anatomical structures to capture certain anatomical regions. MedDiff-FM manages several downstream tasks seamlessly, including image denoising, anomaly detection, and image synthesis. MedDiff-FM is also capable of performing super-resolution, lesion generation, and lesion inpainting by rapidly fine-tuning the diffusion foundation model using ControlNet with task-specific conditions. The experimental results demonstrate the effectiveness of MedDiff-FM in addressing diverse downstream medical image tasks.
\end{abstract}


\begin{figure*}[!t]
\centering
\includegraphics[page=1, trim={1cm, 6.5cm, 1cm, 4cm}, clip, width=0.8\linewidth]{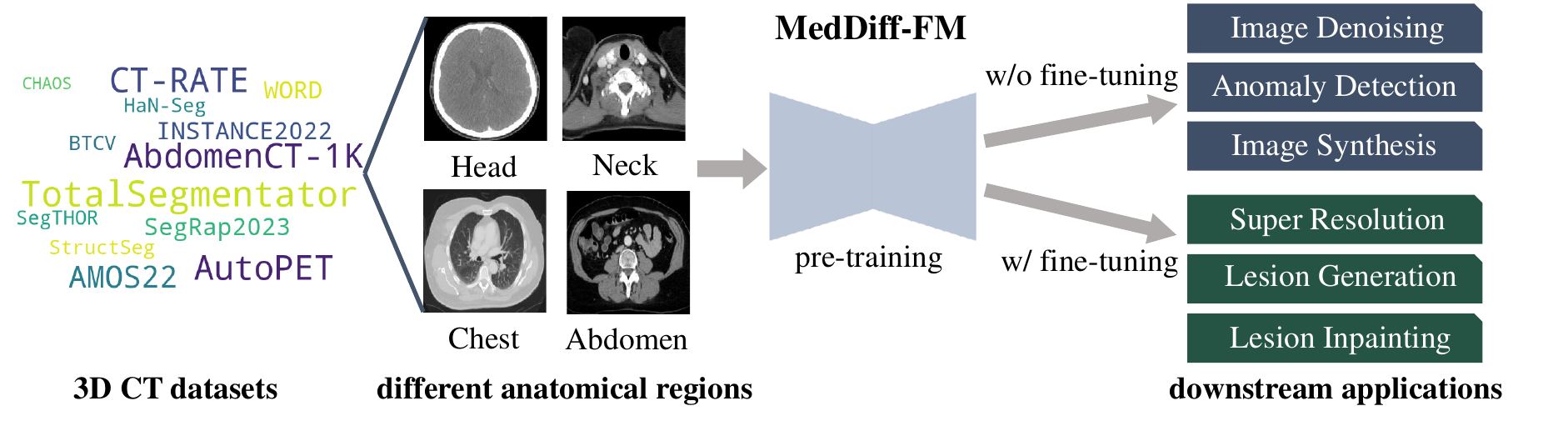}
\caption{An overview of the datasets, anatomical regions, and downstream applications of MedDiff-FM. Covering multiple datasets and diverse anatomical structures, MedDiff-FM supports various downstream tasks with or without fine-tuning.}
\label{fig:overview}
\end{figure*}

\section{Introduction}
\IEEEPARstart{D}{enoising} diffusion probabilistic models (DDPMs)~\cite{ho2020denoising} have gained widespread applications in both natural and medical image domains in recent times. Their ability to generate high-quality and diverse images with stable training dynamics has made DDPMs powerful tools for synthesizing medical image, augmenting datasets, and facilitating segmentation tasks~\cite{chen2024towards, zhang2024diffboost}. Several studies have applied diffusion models to generate both 2D and 3D medical images. For example, ArSDM~\cite{du2023arsdm} and NASDM~\cite{shrivastava2023nasdm} synthesize 2D medical images using semantic masks conditioned semantic diffusion models (SDMs)~\cite{wang2022semantic}, while MedSyn~\cite{xu2024medsyn} and GuideGen~~\cite{dai2024guidegen} are designed to generate 3D CT images using textual and semantic conditions.

The notable success achieved by DDPMs in image synthesis highlights their promising potential beyond generation alone. As a result, diffusion models have been extended to tasks such as image super-resolution~\cite{saharia2022image, yue2024resshift}, and image editing~~\cite{meng2021sdedit, brooks2023instructpix2pix}, further showcasing their versatility. Utilizing pre-trained text-to-image diffusion models for downstream tasks~\cite{zhao2023unleashing, kondapaneni2024text} is also a promising approach. For medical images, diffusion models have been applied to image denoising~\cite{liu2025diffusion}, anomaly detection~\cite{wolleb2022diffusion, wyatt2022anoddpm}, and super-resolution~\cite{zhao2024mri, dong2025flow} tasks.

However, most existing methods focus on specific medical image tasks, particular anatomical regions, and limited datasets, resulting in models that \textbf{fail to generalize across diverse clinical scenarios}. Specifically, these trained diffusion models are typically tailored to specific organs or regions, restricting their applicability to particular anatomical scales or structures. This isolation hinders the sharing of clinical priors and knowledge transfer between models, while also imposing significant computational and maintenance burdens. Moreover, these separate models are task-specific, designed to solve a single task~\cite{liu2025diffusion, wyatt2022anoddpm, dong2025flow} without exploring the intrinsic capabilities of diffusion models or their potential to generalize over a wide range of medical imaging tasks.

In response to these limitations, we propose MedDiff-FM, a pre-trained diffusion foundation model that aims to \textit{cover different anatomical regions, leverage large-scale datasets, and generalize across a variety of medical imaging tasks}. MedDiff-FM deals with multiple anatomical regions, including the head, neck, chest, and abdomen, utilizes publicly available medical image datasets from different institutions, and handles a diverse range of downstream applications. This not only reduces the computational overhead associated with training and maintaining multiple models, but also facilitates the sharing of knowledge across tasks and regions, enhancing the generalization ability of the model. 

To build a unified and scalable generative pre-trained model for CT images, it is important to take into account the variations in image resolution, spacing, and intensity differences across different organs and tissues. MedDiff-FM tackles these challenges by accommodating multi-level medical images, accepting both image-level and patch-level inputs. This multi-level approach allows the model to handle the variations in resolution and spacing that are commonly encountered in CT scans, especially across different anatomical regions. To construct the relationships between multi-level inputs, we draw insights from Patch Diffusion~\cite{wang2024patch}, a patch-level diffusion model that adopts patch coordinate conditioning. We not only adapt the coordinate position conditioning from 2D to 3D for medical images but also advance multi-level relationships. The proposed position embedding plays a critical role in establishing multi-level spatial relationships of 3D CT images. By incorporating position information at both image-level and patch-level, the model is able to capture local textures while also maintaining global consistency across the entire image. Furthermore, MedDiff-FM leverages coarse region class and fine-grained anatomical structure conditions to better capture intensity variations across different organs and tissues, facilitating its ability to generalize across different anatomical regions.

MedDiff-FM handles a diverse range of downstream tasks, including image denoising, anomaly detection, and image synthesis, without the need for fine-tuning. Through fine-tuning with ControlNet, MedDiff-FM is also able to perform super-resolution, lesion generation, and lesion inpainting, under task-specific conditions. When processing entire CT volumes during inference, MedDiff-FM employs a patch-based sliding window sampling strategy with overlapping windows and smoothed noise estimates~\cite{ozdenizci2023restoring} to mitigate boundary artifacts. Experimental results indicate that MedDiff-FM demonstrates superior generalization across diverse tasks and anatomical regions, demonstrating its potential to effectively handle diverse clinical scenarios.

The contributions of this work are summarized as follows:
\begin{itemize}
\item We propose MedDiff-FM, a diffusion-based foundation model that leverages 3D CT images from diverse datasets and multiple anatomical regions, to pre-train a diffusion foundation model for handling a wide range of medical image tasks.
\item MedDiff-FM deals with medical images flexibly, achieving multi-level integrated image processing, and leverages position embedding to build spatial relationships between image-level and patch-level 3D CT images, along with coarse region classes and fine-grained anatomical structures to condition certain anatomical regions.
\item MedDiff-FM provides off-the-shelf applications for several downstream medical tasks, including image denoising, anomaly detection, and image synthesis, representing strong capabilities.
\item Through efficient fine-tuning of MedDiff-FM via ControlNet under task-specific conditions, MedDiff-FM demonstrates effective super-resolution, lesion generation, and lesion inpainting.
\end{itemize}

\begin{figure*}[!t]
\centering
\includegraphics[page=2, trim={0cm, 3cm, 1cm, 1cm}, clip, width=0.75\linewidth]{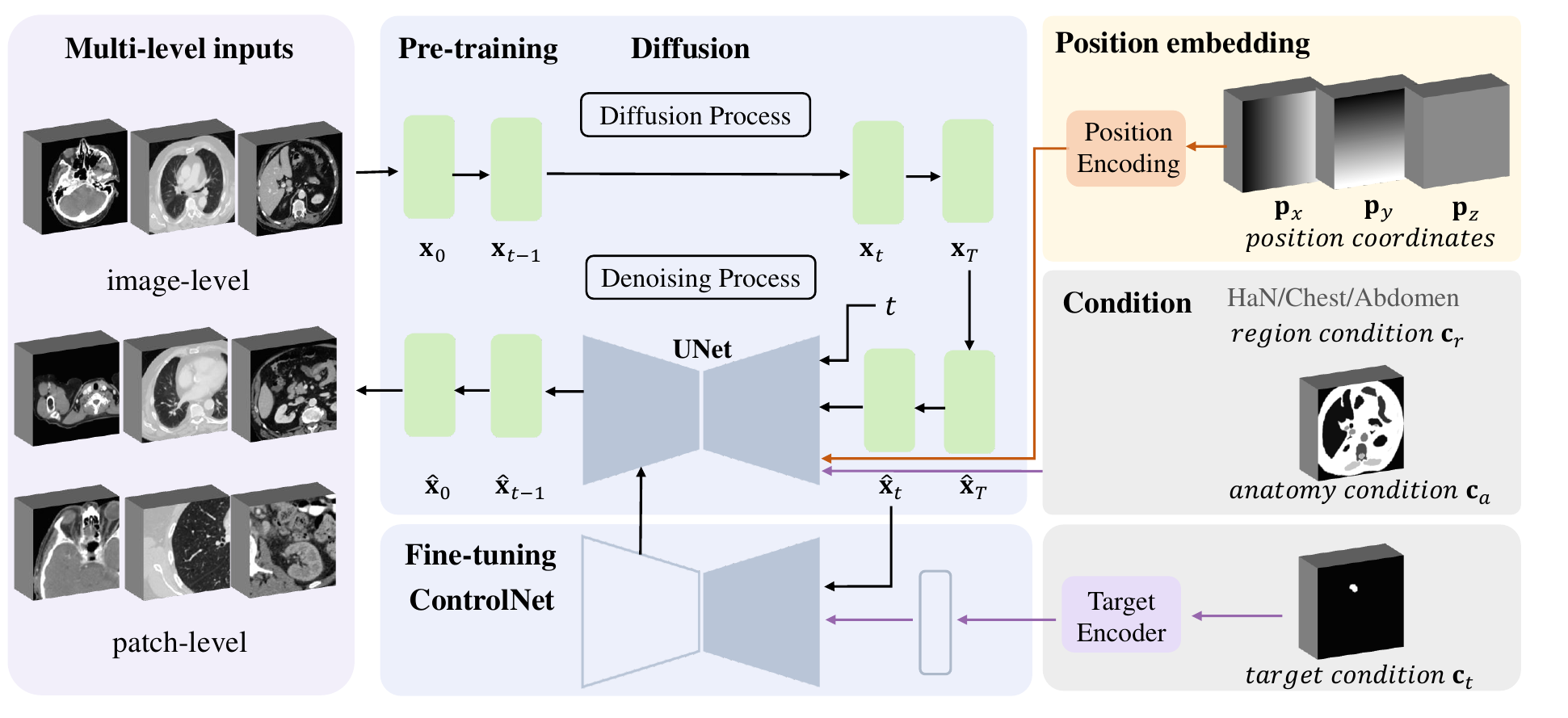}
\caption{The pre-training and fine-tuning pipelines of MedDiff-FM. MedDiff-FM accommodates multi-level medical image inputs to handle the diversity in CT sizes and spacings, leverages positional embeddings to build multi-level spatial relationships, and utilizes both coarse region conditions and fine-grained anatomical conditions. During fine-tuning, task-specific conditions are incorporated via ControlNet.}
\label{fig:pipeline}
\end{figure*}

\section{Related Work}
\subsection{Patch-based Diffusion Models}
Recently, denoising diffusion probabilistic models~\cite{ho2020denoising} have demonstrated superior generative capabilities due to sample quality and diversity, while maintaining training stability. However, in the context of 3D medical images, where voxel dimensions are often large, the patch sizes that diffusion models can process are limited. Although image-level processing captures global information in medical images, it also requires patch-level processing to capture local details.

Patch Diffusion~\cite{wang2024patch} is a patch-level diffusion model training method, which adopts patch coordinate conditioning and patch size scheduling to balance global encoding effectiveness and training efficiency.
To deal with multi-level medical images and construct the relationship between image-level and patch-level inputs, MedDiff-FM not only utilizes the coordinate position conditioning from Patch Diffusion but also advances multi-level representations, and extends the coordinate position conditioning from 2D to 3D, in order to establish the multi-level relationships of medical images.

Özdenizci \emph{et al.}~\cite{ozdenizci2023restoring} design a patch-based diffusion approach that processes images of arbitrary size during inference and utilizes smoothed noise estimates across overlapping patches.
Therefore, to deal with entire CT volumes with the patch-level diffusion model, MedDiff-FM leverages patch-based sliding window sampling strategy with overlapping windows and smoothed noise estimates to eliminate artificial boundaries.

\subsection{Diffusion Applications in Natural Images}
In the natural image domain, stable diffusion~\cite{rombach2022high} has emerged as a powerful pre-trained text-to-image generation model. Stable diffusion is a latent diffusion model~\cite{rombach2022high} that performs the diffusion process in the low-dimensional latent space instead of the high-dimensional pixel space. ControlNet~\cite{zhang2023adding} generalizes the pre-trained text-to-image diffusion models to diverse conditions, such as semantic maps~\cite{wang2022semantic, singh2023high}, canny edges, and depth maps. In addition to text-to-image diffusion models, text-to-video diffusion models~\cite{blattmann2023align, brooks2024video} have also witnessed significant development, such as Sora~\cite{brooks2024video}.

In addition to unconditional and conditional image generation tasks, diffusion models have also been applied to other tasks~\cite{croitoru2023diffusion}. For example, image super-resolution~\cite{saharia2022image, yue2024resshift}, image editing~\cite{meng2021sdedit, brooks2023instructpix2pix}, and image-to-image translation~\cite{saharia2022palette, tumanyan2023plug}. Moreover, Zhao \emph{et al.}~\cite{zhao2023unleashing} and Kondapaneni \emph{et al.}~\cite{kondapaneni2024text} adapt pre-trained text-to-image diffusion models to diverse downstream tasks, and show the capabilities of pre-trained diffusion models.

\subsection{Diffusion Applications for Medical Images}
Beyond the natural image domain, diffusion models are also flourishing in the medical image domain, including 2D and 3D medical image modalities.
ArSDM~\cite{du2023arsdm} generates colonoscopy polyp images with polyp masks.
NASDM~\cite{shrivastava2023nasdm} synthesizes nuclei pathology images conditioned on nuclei masks.
Medical Diffusion~\cite{khader2023denoising} generates unconditional 3D CT and MRI images.
Zhuang \emph{et al.}~\cite{zhuang2023semantic} propose to generate abdominal CT images using semantic masks.
MedSyn~\cite{xu2024medsyn} is proposed for generating high-fidelity 3D chest CT images with textual guidance.
GuideGen~\cite{dai2024guidegen} synthesizes paired anatomical structures and CT scans via text-guided diffusion models.
MAISI~\cite{guo2025maisi} is developed for producing high-resolution 3D CT volumes across different body regions.
Med-DDPM~\cite{dorjsembe2024conditional} is introduced to generate 3D brain MRIs with brain tumors.
DiffTumor~\cite{chen2024towards} aims to synthesize tumors on healthy abdominal organs using diffusion models.

Diffusion models for medical images have diverse applications under different medical image scenarios~\cite{kazerouni2023diffusion} beyond image synthesis. For example,
Dn-Dp~\cite{liu2025diffusion} achieves zero-shot low-dose CT denoising by training solely on normal-dose CT images.
Wolleb \emph{et al.}~\cite{wolleb2022diffusion} utilize a class-conditional diffusion model for the anomaly detection of brain tumors and pleural effusion.
AnoDDPM~\cite{wyatt2022anoddpm} is designed for the anomaly detection of brain tumors from 2D MRI images.
Jimenez-Perez \emph{et al.}~\cite{jimenez2024dino} attempt to pre-train a diffusion model on chest X-rays for reconstruction and segmentation tasks.

\section{Methodology}
In this section, we first provide an overview of MedDiff-FM, then introduce the multi-level integrated medical diffusion foundation model and the position embedding for 3D CT images, and finally discuss the application of MedDiff-FM to downstream tasks.

\subsection{Overview of MedDiff-FM}
Fig.~\ref{fig:overview} illustrates the overview of MedDiff-FM. MedDiff-FM is trained on a diverse range of publicly available 3D CT datasets from different institutions. These 3D CT datasets cover different anatomical regions, including the head, neck, chest, and abdomen, encompassing a variety of anatomical structures and eliminating the limitations of focusing on a single anatomical region. The pre-training of MedDiff-FM on different anatomical regions and diverse medical image datasets enables it to be applied to downstream tasks across multiple anatomical regions, demonstrating strong generalization capabilities. MedDiff-FM can be directly applied to downstream tasks, including image denoising, anomaly detection, and image synthesis, without the need for fine-tuning, which significantly conserves resources and enhances convenience. Additionally, by fine-tuning MedDiff-FM, super-resolution, lesion generation, and lesion inpainting can be achieved by incorporating task-specific conditions using ControlNet.

\subsection{Multi-level Integrated Medical Diffusion Foundation Model}
As depicted in Fig.~\ref{fig:pipeline}, the diffusion foundation model accepts multi-level medical image inputs, specifically image-level inputs and patch-level inputs, in order to accommodate the varying resolutions and spacings of CT images. Given a 3D medical image $\mathbf{x}$, we randomly apply one of three operations with equal probability: resizing the image to the patch size, randomly cropping the image to twice the patch size and then resizing it to the patch size, or randomly cropping the image to the patch size. The first operation yields image-level inputs, whereas the latter two yield patch-level inputs. The patch size used in this paper is 128 × 128 × 128. The multi-level input $\mathbf{x}_0$ follows the data distribution $q(\mathbf{x})$. As shown in~\eqref{eq:q}, for $T$ diffusion timesteps with variances $\beta_1,...,\beta_T$, it produces a sequence of noisy images $\mathbf{x}_1,...,\mathbf{x}_T$, where $\mathbf{x}_T$ follows a standard Gaussian distribution. In the denoising process, the unconditional denoising U-Net~\cite{ronneberger2015u} progressively transforms random noises into images via~\eqref{eq:p}. In contrast, our conditional denoising U-Net takes as input the noisy image $\mathbf{\hat{x}}_t$, the current timestep $t$, the position embedding, and additional conditions, and outputs the estimated noise $\boldsymbol{\hat{\epsilon}}$.

\begin{equation}
q(\mathbf{x}_t|\mathbf{x}_{t-1})=\mathcal{N}(\mathbf{x}_t;\sqrt{1-\beta_t}\mathbf{x}_{t-1},\beta_t\mathbf{I})
\label{eq:q}
\end{equation}

\begin{equation}
p_\theta(\mathbf{x}_{t-1}|\mathbf{x}_t)=\mathcal{N}(\mathbf{x}_{t-1};\boldsymbol{\mu}_\theta(\mathbf{x}_t,t),\boldsymbol{\Sigma}_\theta(\mathbf{x}_t,t))
\label{eq:p}
\end{equation}

During the pre-training of the diffusion foundation model, we leverage two conditioning signals, the coarse region condition $\mathbf{c}_r$ and the fine-grained anatomy condition $\mathbf{c}_a$, to achieve multi-perspective control over the denoising process and enable the model to better capture the intensity differences across different organs and tissues. The region condition $\mathbf{c}_r$ utilizes anatomical region classes that indicate whether the anatomical region belongs to the head and neck (HaN), chest, or abdomen. The anatomy condition $\mathbf{c}_a$ leverages anatomical structure masks derived from TotalSegmentator~\cite{wasserthal2023totalsegmentator} and thresholding methods to enforce anatomical control via channel-wise concatenation. Since the diffusion foundation model takes multi-level medical image inputs, we aim to establish explicit spatial relationships between image-level and patch-level inputs. We adopt position encoding to obtain position embedding $\mathbf{p}_e$ of the X, Y, and Z coordinates, which we will discuss further in Section~\ref{sec:relationship}.

The overall training objective of MedDiff-FM is formulated as:
\begin{equation}
\mathbb{E}_{\mathbf{x}_0,\boldsymbol{\epsilon},t,\mathbf{p}_e,\mathbf{c}_r,\mathbf{c}_a}\left[||\boldsymbol{\epsilon}-\boldsymbol{\epsilon}_\theta(\mathbf{x}_t,t,\mathbf{p}_e,\mathbf{c}_r,\mathbf{c}_a)||\right].
\label{eq:loss1}
\end{equation}

When it is necessary to fine-tune the diffusion foundation model, we utilize ControlNet to incorporate task-specific conditions. ControlNet leverages the weights of the neural network blocks, creating a locked copy to preserve the knowledge of the pre-trained diffusion model and a trainable copy to adapt to additional conditions. These copies are connected through zero convolution modules which gradually adjust the parameters starting from zero. A target encoder is employed to extract features from the target condition $\mathbf{c}_t$, enabling a more effective injection of auxiliary conditions into the diffusion foundation model.

The objective function for fine-tuning MedDiff-FM is formulated as:
\begin{equation}
\mathbb{E}_{\mathbf{x}_0,\boldsymbol{\epsilon},t,\mathbf{p}_e,\mathbf{c}_r,\mathbf{c}_a,\mathbf{c}_t}\left[||\boldsymbol{\epsilon}-\boldsymbol{\epsilon}_\theta(\mathbf{x}_t,t,\mathbf{p}_e,\mathbf{c}_r,\mathbf{c}_a,\mathbf{c}_t)||\right].
\label{eq:loss2}
\end{equation}

\subsection{3D CT Image Position Embedding}\label{sec:relationship}
The voxel dimensions in 3D medical images are often large; however, the image sizes that diffusion models can process are limited. The image-level processing may lead to the neglect of local details, while the patch-level processing may lose holistic perception. Therefore, we propose the diffusion foundation model that is designed to handle both image-level and patch-level inputs simultaneously, aiming to integrate multi-level medical image information.

\begin{figure}[htbp]
\centering
\includegraphics[page=3, trim={6cm, 9cm, 9cm, 5cm}, clip, width=0.85\linewidth]{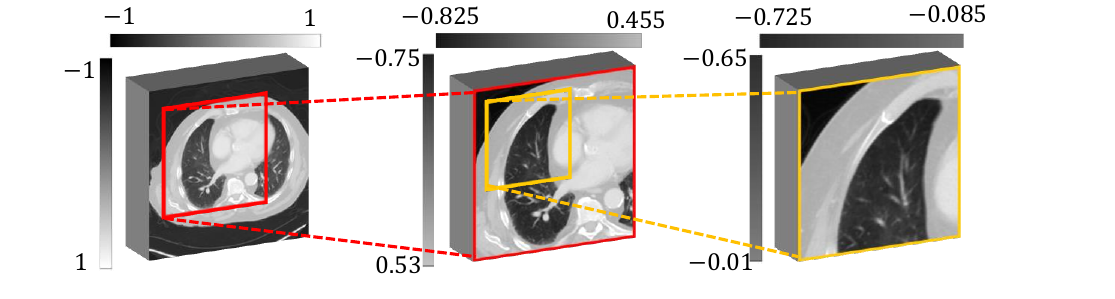}
\caption{The multi-level position relationships constructed based on position embedding. Positional relationships illustrated in the X–Y plane, with the Z-axis handled similarly.}
\label{fig:position}
\end{figure}

To construct relationships between image-level and patch-level inputs, we draw insights from Patch Diffusion~\cite{wang2024patch}, a patch-level diffusion model that utilizes patch coordinate conditioning. Instead of solely using patch coordinates to represent the relative position of the patch to the original image, we construct multi-level position relationships, as illustrated in Fig.~\ref{fig:position}. The multi-level position relationships establish explicit connections between the global and local information in medical images. In addition, we adapt the 2D coordinate position conditioning to 3D for medical images. The coordinate positions are pixel-level and normalized to $[-1,1]$, where $(-1,-1,-1)$ denotes the upper left back corner and $(1,1,1)$ denotes the bottom right front corner. The three position coordinate channels $\mathbf{p}_x$, $\mathbf{p}_y$, and $\mathbf{p}_z$ represent the X, Y, and Z coordinate positions, respectively. Furthermore, we employ the position encoding function $\text{PE}(\cdot)$~\cite{mildenhall2021nerf} to better encode positional information, as formulated in~\eqref{eq:position}, with $L$ denoting the maximum frequency. The obtained position encoding $\mathbf{p}_e$ is then concatenated with the input image and incorporated into the model.

\begin{align}
\mathbf{p}_c &= \text{concat}(\mathbf{p}_x,\mathbf{p}_y,\mathbf{p}_z) \\
\mathbf{p}_e &= \text{PE}(\mathbf{p}_c) \nonumber \\
&=\big(\sin(2^0\pi\mathbf{p}_c),\,\cos(2^0\pi\mathbf{p}_c), \nonumber \\
&\quad\ldots, \nonumber \\
&\quad\sin(2^{L-1}\pi\mathbf{p}_c),\,\cos(2^{L-1}\pi\mathbf{p}_c)\big)
\label{eq:position}
\end{align}

\begin{figure}[htbp]
\centering
\includegraphics[page=4, trim={5cm, 3.5cm, 5cm, 2cm}, clip, width=0.9\linewidth]{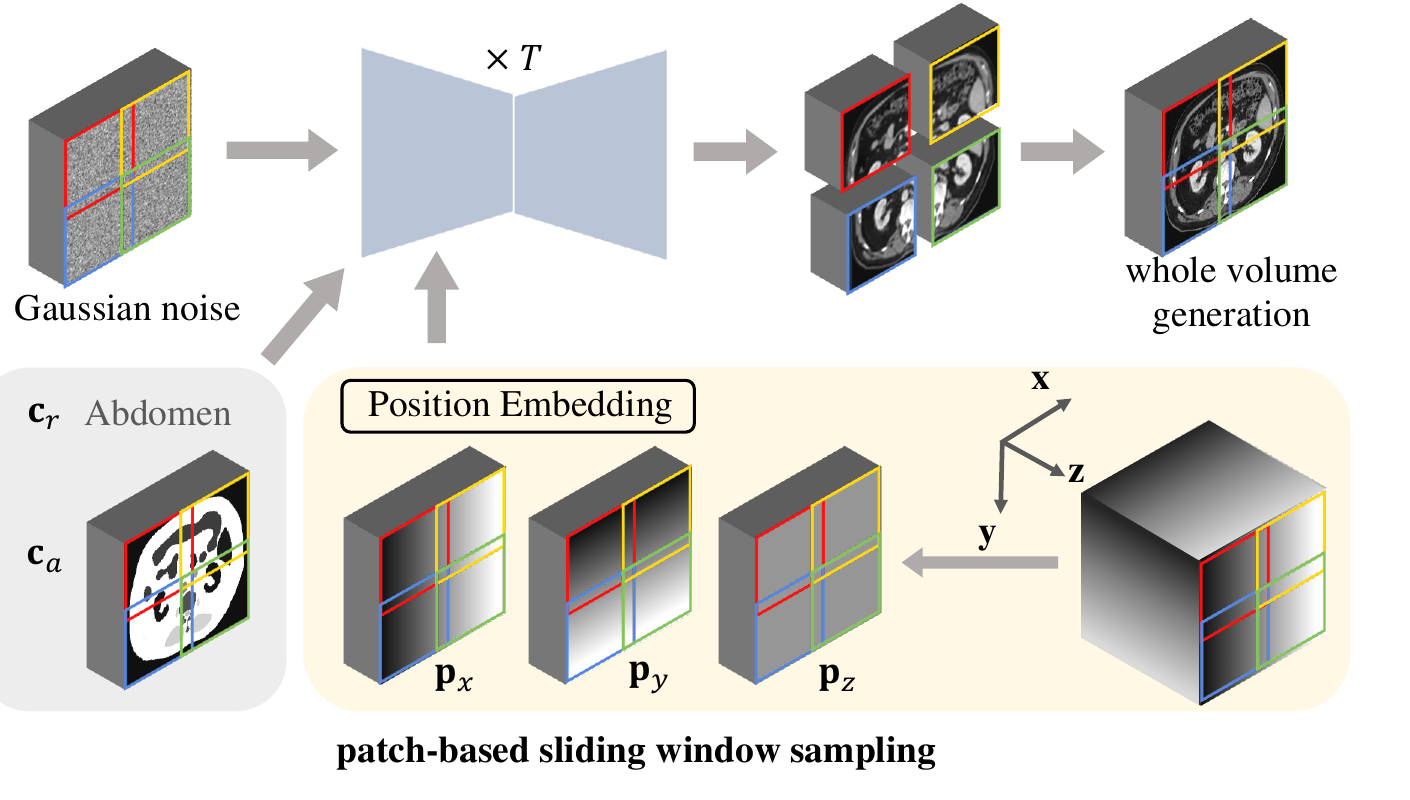}
\caption{The process of patch-level whole-volume synthesis. MedDiff-FM employs a patch-based sliding window sampling strategy, using overlapping windows and smoothed noise estimates to effectively eliminate boundary artifacts.}
\label{fig:method}
\end{figure}

During inference, leveraging the established multi-level position relationships, MedDiff-FM utilizes a patch-based sliding window sampling strategy to deal with entire CT volume processing or generation using the patch-level diffusion model. The multi-level position relationships establish spatial relations between each patch and the whole volume. An example of patch-level whole volume synthesis is shown in Fig.~\ref{fig:method}. The diffusion model generates patch-level images, which are then combined to form the entire CT volume. To mitigate artificial boundaries between overlapping windows, we employ smoothed noise estimates~\cite{ozdenizci2023restoring} across overlapping patches. At each denoising timestep $t$, the mean estimated noise based sampling updates are applied to overlapping pixels across patches.

\subsection{Downstream Tasks without Fine-tuning}
MedDiff-FM can be used to accomplish multiple downstream tasks, including image synthesis, image denoising, and anomaly detection, without the need for fine-tuning. Image synthesis is beneficial for augmenting medical image datasets and enhancing data diversity, as shown in Fig.~\ref{fig:method}, MedDiff-FM synthesizes whole CT volumes of flexible sizes based on patch-based sliding window sampling.

\subsubsection{Image Denoising}\label{sec:imagedenoise}
CT is a prevalent imaging modality in clinical diagnosis, but the associated radiation exposure poses potential health risks. Low-dose CT (LDCT) effectively reduces radiation dose compared to full-dose CT (FDCT) but often suffers from significant noise and artifacts, which can impair diagnostic accuracy. Consequently, image denoising techniques are essential for enhancing the quality of LDCT images. The inverse problem aims to recover unknown signals from observed measurements, and the corruption process is formulated as~\eqref{eq:inverseproblem}, where $\mathcal{H}$ denotes a degradation operator and $\mathbf{n}$ represents additive noise. Image denoising is an inverse problem to recover the clean image $\mathbf{x}_0$ from the noisy image $\mathbf{y}_0$, and $\mathcal{H}(\mathbf{x}_0)=\mathbf{x}_0$ in the case of image denoising. The solution of inverse problems can be formulated as~\eqref{eq:inversesolution}, where the first term is the data term with $\sigma_n$ denoting the standard deviation of the additive noise, and the second term is the prior term with the regularization parameter $\lambda$. The data term guarantees that the solution is consistent with the degradation process, while the prior term encourages the solution to follow the desired data distribution.

\begin{equation}
\mathbf{y}_0=\mathcal{H}(\mathbf{x}_0)+\mathbf{n}
\label{eq:inverseproblem}
\end{equation}

\begin{equation}
\hat{\mathbf{x}}_0=\arg\min_{\mathbf{x}}\frac{1}{2\sigma_n^2}\left\|\mathbf{y}_0-\mathcal{H}(\mathbf{x})\right\|^2+\lambda\mathcal{R}(\mathbf{x})
\label{eq:inversesolution}
\end{equation}

Leveraging a pre-trained diffusion model for plug-and-play image denoising, we extend the sampling approach of DiffPIR~\cite{zhu2023denoising} to image denoising task. DiffPIR is proposed to address plug-and-play image restoration for inverse problems such as super-resolution and image deblurring. It decouples the data term and the prior term in~\eqref{eq:inversesolution}, and solves the decoupled subproblems iteratively. At each step of the reverse diffusion sampling, DiffPIR employs a diffusion model as a generative denoiser prior and then solves the data proximal subproblem. We adapt DiffPIR to the image denoising task, leveraging the pre-trained MedDiff-FM as the denoiser prior as in~\eqref{eq:xrecon}. The analytic solution of the data proximal subproblem for image denoising is~\eqref{eq:xreconhat}, where $\rho_t$ is a trade-off coefficient defined as $\rho_t\triangleq\lambda\sigma_n^2/\bar{\sigma}_t^2$ and $\bar{\sigma}_t=\sqrt{(1-\bar{\alpha}_t)/\bar{\alpha}_t}$.

\begin{equation}
\mathbf{x}_0^{(t)}=\frac{1}{\sqrt{\bar{\alpha}_t}}\left(\mathbf{x}_t-\sqrt{1-\bar{\alpha}_t}\cdot\boldsymbol{\epsilon}_\theta\right)
\label{eq:xrecon}
\end{equation}

\begin{equation}
\hat{\mathbf{x}}_0^{(t)}=\frac{\mathbf{y}_0+\rho_t\cdot\mathbf{x}_0^{(t)}}{1+\rho_t}
\label{eq:xreconhat}
\end{equation}

Since LDCT images are not pure Gaussian noise, we start the reverse sampling process from an intermediate step $t_{start}$. Let $\sigma_n=\bar{\sigma}_t=\sqrt{(1-\bar{\alpha}_t)/\bar{\alpha}_t}$, and by substituting the estimated value of $\sigma_n$, the corresponding approximate $t$ is taken as $t_{start}$. We then derive the expression of $\mathbf{x}_{start}$ corresponding to $t_{start}$. Given $\mathbf{y}_0=\mathbf{x}_0+\sigma_n\mathbf{z}$ and the diffusion process $\mathbf{x}_t=\sqrt{\bar{\alpha}_t}\mathbf{x}_0+\sqrt{1-\bar{\alpha}_t}\mathbf{z}$, we derive the relation in~\eqref{eq:relation} and thus obtain $\mathbf{x}_t=\sqrt{\bar{\alpha}_t}\mathbf{y}_0$. Consequently, the reverse sampling process can be initialized with $\mathbf{x}_{start}=\sqrt{\bar{\alpha}_{start}}\mathbf{y}_0$.

\begin{equation}
\sqrt{\bar{\alpha}_t}\mathbf{y}_0=\sqrt{\bar{\alpha}_t}\mathbf{x}_0+\sqrt{\bar{\alpha}_t}\sigma_n\mathbf{z}=\sqrt{\bar{\alpha}_t}\mathbf{x}_0+\sqrt{1-\bar{\alpha}_t}\mathbf{z}
\label{eq:relation}
\end{equation}

\subsubsection{Anomaly Detection}\label{sec:anomalydetect}
Anomaly detection aims at identifying irregularities or abnormalities in medical images, aiding in disease detection and diagnosis. A large quantity of data is utilized during the pre-training of MedDiff-FM. We consider that the majority of the data contain healthy organs or tissues, and although a small fraction exhibit abnormalities in certain anatomical regions, these regions are normal in most cases. Consequently, MedDiff-FM learns to synthesize normal organs or tissues. Furthermore, when using diffusion models for anomaly detection, it is more important to add noise that corrupts the abnormal regions themselves during inference. To perform anomaly detection on specific anatomical regions, we utilize anatomical structure masks to focus on the regions of interest while masking out the others. Therefore, given the masked abnormal image $\mathbf{\tilde{x}}_0$, we add noise at a fixed time step $t$ to obtain the corresponding noisy abnormal image $\mathbf{\tilde{x}}_t$. We then directly predict the original image, yielding a reconstructed healthy image $\mathbf{\tilde{x}}'_0$. The anomaly map is calculated as $\mathbf{Ano_{map}}=\mathbf{\tilde{x}}_0-\mathbf{\tilde{x}}'_0$. Finally, we use a threshold to binarize the anomaly map and obtain a binary anomaly mask $\mathbf{Ano_{mask}}$.

\subsection{Downstream Tasks with Fine-tuning}
\subsubsection{Super-resolution}\label{sec:superres}
In clinical practice, thick-slice CT is widely used due to its lower cost, fast acquisition, and reduced radiation exposure compared to thin-slice CT. However, its coarse spatial resolution often limits accurate diagnosis. Volumetric super-resolution offers a feasible solution by upsampling in the depth dimension, recovering high-resolution (HR) volumes from low-resolution (LR) volumes along the z-axis. Super-resolution is also an inverse problem to recover the HR image $\mathbf{x}_0$ from the LR image $\mathbf{y}_0$ as represented in~\eqref{eq:inverseproblem}, and $\mathcal{H}$ denotes the downsampling operator with downsampling factor $sf$.

Since pre-trained MedDiff-FM integrates multi-level medical image inputs, it requires fine-tuning for super-resolution tasks. To this end, we use real-paired thick-CTs (LR images) and thin-CTs (HR images) to fine-tune the model via ControlNet, and the task-specific target condition $\mathbf{c}_t$ in~\eqref{eq:loss2} corresponds to the LR images. We adopt DiffPIR~\cite{zhu2023denoising} for plug-and-play volumetric super-resolution, where the data term and the prior term in the solution~\eqref{eq:inversesolution} are decoupled, and the decoupled subproblems are solved iteratively. The fine-tuned MedDiff-FM serves as the denoiser prior as described in~\eqref{eq:xrecon}. And the analytic solution of the data proximal subproblem for super-resolution is formulated as~\eqref{eq:xreconhatsr}, where $\mathcal{H}^\top$ denotes the upsampling operator with upsampling factor $sf$.

\begin{equation}
\hat{\mathbf{x}}_0^{(t)}=\mathbf{x}_0^{(t)}+\frac{\mathcal{H}^\top\left(\mathbf{y}_0-\mathcal{H}(\mathbf{x}_0^{(t)})\right)}{1+\rho_t}
\label{eq:xreconhatsr}
\end{equation}

\subsubsection{Lesion Generation}
Lesion generation is intended to synthesize images containing lesions, which can augment medical datasets and facilitate downstream tasks such as segmentation~\cite{zhang2024diffboost}. By rapidly fine-tuning MedDiff-FM with ControlNet on lesion datasets, high-quality lesion generation can be achieved even with limited training data. During fine-tuning, the lesion mask serves as the target condition $\mathbf{c}_t$, as in~\eqref{eq:loss2}, guiding the generation process of lesion images.

\subsubsection{Lesion Inpainting}
Lesion inpainting aims to synthesize lesions in normal medical images given lesion masks, enabling realistic and controllable lesion generation for data augmentation~\cite{chen2024towards}. We directly leverage the fine-tuned lesion generation model to perform lesion inpainting rather than repeatedly fine-tuning. Since the lesion generation model is already capable of generating lesions, it can be generalized to the lesion inpainting task straightforwardly. Lesion inpainting can better leverage the information from the original CT image rather than relying entirely on synthesis. We adopt the image inpainting method proposed in RePaint~\cite{lugmayr2022repaint} as described in~\eqref{eq:repaint}, where $\mathbf{c}_t$ denotes the lesion mask. During inference, this method combines the known region from the original image $\mathbf{x}^{\text{known}}_{t-1}$ obtained from~\eqref{eq:q} with the inpainted region $\mathbf{x}^{\text{unknown}}_{t-1}$ generated by the reverse diffusion process at each time step.

\begin{equation}
\mathbf{x}^{\text{inpaint}}_{t-1}=(1-\mathbf{c}_t)\odot\mathbf{x}^{\text{known}}_{t-1}+\mathbf{c}_t\odot\mathbf{x}^{\text{unknown}}_{t-1}
\label{eq:repaint}
\end{equation}

\section{Experiments and Results}
In this section, we first describe the experimental setup, including datasets, implementation details, and evaluation metrics. Next, we evaluate the downstream tasks, which can be divided into two categories: those that do not require fine-tuning of MedDiff-FM and those that require fine-tuning of MedDiff-FM. The first category includes image synthesis, image denoising, and anomaly detection, while the second category includes super-resolution, lesion generation, and lesion inpainting.

\begin{table}[htbp]
\centering
\footnotesize
\caption{CT datasets for the development of MedDiff-FM.}
\begin{tabular}{clc}
\toprule
Region & Dataset Name & \# Cases \\
\midrule
\multirow{4}{*}{HaN}
& StructSeg~\cite{h75x-gt46-23} & 50 \\
& INSTANCE2022~\cite{li2023state, li2021hematoma} & 130 \\
& HaN-Seg~\cite{podobnik2023han} & 42 \\
& SegRap2023~\cite{luo2023segrap2023} & 140 \\
\midrule
\multirow{2}{*}{Chest}
& SegTHOR~\cite{lambert2020segthor} & 40 \\
& CT-RATE~\cite{hamamci2024foundation} & 1,000 \\
\midrule
\multirow{5}{*}{Abdomen}
& AbdomenCT-1K~\cite{ma2021abdomenct} & 1,062 \\
& AMOS22~\cite{ji2022amos} & 500 \\
& BTCV~\cite{landman2015miccai} & 30 \\
& CHAOS~\cite{CHAOS2021, CHAOSdata2019} & 20 \\ 
& WORD~\cite{luo2021word} & 120 \\
\midrule
\multirow{2}{*}{Whole Body}
& TotalSegmentator~\cite{wasserthal2023totalsegmentator} & 1,228 \\
& AutoPET~\cite{gatidis2022whole} & 1,014 \\
\midrule
\textbf{Total} & & \textbf{5,376} \\
\bottomrule
\end{tabular}
\label{tab:dataset}
\end{table}

\begin{table}[htbp]
\centering
\footnotesize
\caption{CT datasets for downstream tasks.}
\begin{tabular}{clc}
\toprule
Task & Dataset Name & \# Cases \\
\midrule
Image Denoising & Mayo 2016~\cite{mccollough2017low} & 10 \\
\midrule
Anomaly Detection & BMAD Liver~\cite{bao2024bmad} & 3,201 (slices) \\
\midrule
Super-Resolution & RPLHR-CT~\cite{yu2022rplhr} & 250 \\
\midrule
Lesion Generation & MSD-Lung~\cite{simpson2019large, antonelli2022medical} & 63 \\
\& & MED-LN~\cite{roth2014new} & 90 \\
Lesion Inpainting & MSD-Liver~\cite{simpson2019large, antonelli2022medical} & 131 \\
& ABD-LN~\cite{roth2014new} & 86 \\
\bottomrule
\end{tabular}
\label{tab:taskdataset}
\end{table}

\begin{table*}[htbp]
\centering
\footnotesize
\caption{Ablation study on the effectiveness of components in MedDiff-FM. Each component is incrementally added to the patch-level DDPM baseline.}
\begin{tabular}{clccccc}
\toprule
Anatomical Region & Method & MS-SSIM $\uparrow$ & LPIPS $\downarrow$ & FID $\downarrow$ & MMD $\downarrow$ & Dice $\uparrow$ \\
\midrule
\multirow{4}{*}{HaN}
& patch-level baseline       & 0.7227 & 0.0185 & 0.1973 & 0.0513 & 0.7491 \\
& + multi-level integration  & 0.7867 & 0.0125 & 0.1564 & 0.0301 & 0.8843 \\
& + position                 & 0.8049 & 0.0132 & 0.1101 & 0.0293 & 0.8923 \\
& + position embedding       & 0.8034 & 0.0110 & 0.1074 & 0.0275 & 0.8949 \\
\midrule
\multirow{4}{*}{Chest}
& patch-level baseline       & 0.7799 & 0.0175 & 0.3121 & 0.1219 & 0.4909 \\
& + multi-level integration  & 0.8344 & 0.0116 & 0.1961 & 0.0753 & 0.6372 \\
& + position                 & 0.8443 & 0.0121 & 0.1069 & 0.0652 & 0.6997 \\
& + position embedding       & 0.8480 & 0.0130 & 0.0749 & 0.0648 & 0.7076 \\
\midrule
\multirow{4}{*}{Abdomen}
& patch-level baseline       & 0.6009 & 0.0242 & 0.2792 & 0.0489 & 0.6998 \\
& + multi-level integration  & 0.7256 & 0.0148 & 0.2130 & 0.0325 & 0.8318 \\
& + position                 & 0.7255 & 0.0157 & 0.2402 & 0.0390 & 0.8503 \\
& + position embedding       & 0.7439 & 0.0136 & 0.2595 & 0.0303 & 0.8524 \\
\midrule
\multirow{4}{*}{Overall}
& patch-level baseline       & 0.7011 & 0.0201 & 0.1096 & 0.0740 & 0.6466 \\
& + multi-level integration  & 0.7823 & 0.0130 & 0.0819 & 0.0460 & 0.7844 \\
& + position                 & 0.7916 & 0.0137 & 0.0667 & 0.0445 & 0.8141 \\
& + position embedding       & 0.7984 & 0.0125 & 0.0655 & 0.0409 & 0.8183 \\
\bottomrule
\end{tabular}
\label{tab:ablation}
\end{table*}

\subsection{Experimental Setup}
\subsubsection{Datasets}
\textbf{MedDiff-FM Datasets.}
We collect several publicly available medical image datasets, covering diverse anatomical regions and structures, for developing MedDiff-FM. The medical image datasets and their corresponding number of cases used in our experiments are shown in Table~\ref{tab:dataset}. We gather a total of 5,376 CT volumes, consisting of 362 head and neck volumes, 1,040 chest volumes, 1,732 abdomen volumes, and 2,242 whole body volumes. To balance the number of samples across different anatomical regions, we only adopt 1,000 chest cases from the CT-RATE~\cite{hamamci2024foundation} dataset. These datasets are randomly divided into 90\% for MedDiff-FM training and 5\% for validation, with the remaining 5\% reserved for the evaluation of MedDiff-FM on the downstream image synthesis task.

\textbf{Task-specific Datasets.} 
To further validate MedDiff-FM for other downstream tasks, we adopt the datasets listed in Table~\ref{tab:taskdataset}. The Mayo 2016 dataset~\cite{mccollough2017low}, also known as the 2016 NIH-AAPM-Mayo Clinic Low-Dose CT Grand Challenge, contains 1\,mm full-dose and quarter-dose CT images from 10 patients. Covering both the chest and abdomen regions, the Mayo 2016 dataset is used to evaluate the image denoising capabilities of MedDiff-FM. The Liver CT benchmark from BMAD~\cite{bao2024bmad}, benchmarks for medical anomaly detection, is used for anomaly detection evaluation. We utilize the real-paired RPLHR-CT~\cite{yu2022rplhr} dataset, which contains real-world thick-CTs (5\,mm slice thickness) and thin-CTs (1\,mm slice thickness), to assess super-resolution performance. Additionally, we adopt four datasets to evaluate lesion generation and lesion inpainting tasks. Specifically, MSD-Lung and MSD-Liver are Task06 and Task03 of Medical Segmentation Decathlon (MSD)~\cite{simpson2019large, antonelli2022medical}. MED-LN and ABD-LN are acquired from~\cite{roth2014new}, containing mediastinal and abdominal lymph nodes, respectively. The four datasets are each randomly divided, with 80\% used for fine-tuning MedDiff-FM, specifically for ControlNet training and validation, and the remaining 20\% for testing.

\subsubsection{Implementation Details}
We implement all the methods using PyTorch and carry out all the experiments using NVIDIA GeForce RTX 3090 GPUs.
To obtain the anatomical structures used in MedDiff-FM, we leverage the automated whole body medical image segmentation tool, TotalSegmentator v2~\cite{wasserthal2023totalsegmentator}, to segment CT images. Incorporating the 117 classes segmented by TotalSegmentator, we further derive the body class using a thresholding method, resulting in 118 classes in total.
For data preprocessing, we first resample the image voxel spacing to 1.0\,mm × 1.0\,mm × 1.0\,mm. Next, for whole body CT images, we split the head and neck, chest, and abdomen regions based on the segmentation results from TotalSegmentator and crop the images corresponding to these anatomical regions. To differentiate and extract features across different anatomical regions, we apply varying window widths and levels for window truncation to adapt to each region. For the head and neck region, the window level is set to 50 and the window width to 400; for the chest region, the window level is -500 and the window width is 1800; and for the abdomen region, the window level is 60 and the window width is 360. Subsequently, the data ranges are normalized to $[-1,1]$.
The diffusion timesteps $T$ is 1,000 with cosine noise schedule. The patch size used in our experiment is 128 × 128 × 128. The model channels for MedDiff-FM and ControlNet are 32 and the number of residual blocks is 1, with the spatial transformer operating at a spatial resolution of 16 × 16 × 16. For training, we utilize $L_1$ loss and Adam optimizer, with a learning rate of $10^{-4}$, a batch size of 1, and 4 gradient accumulation steps. The MedDiff-FM is trained for around 150 epochs, and the ControlNet is trained for around 10k steps.

\begin{table*}[htbp]
\centering
\footnotesize
\caption{Comparison with state-of-the-art methods on conditional CT volume synthesis.}
\begin{tabular}{clccccc}
\toprule
Anatomical Region & Method & MS-SSIM $\uparrow$ & LPIPS $\downarrow$ & FID $\downarrow$ & MMD $\downarrow$ & Dice $\uparrow$ \\
\midrule
\multirow{3}{*}{HaN}
& DDPM        & 0.7227 & 0.0185 & 0.1973 & 0.0513 & 0.7491 \\
& MAISI       & 0.6963 & 0.0150 & 0.1572 & 0.0414 & 0.8164 \\
& MedDiff-FM  & \textbf{0.8034} & \textbf{0.0110} & \textbf{0.1074} & \textbf{0.0275} & \textbf{0.8949} \\
\midrule
\multirow{3}{*}{Chest}
& DDPM        & 0.7799 & 0.0175 & 0.3121 & 0.1219 & 0.4909 \\
& MAISI       & 0.7063 & 0.0240 & 0.1731 & \textbf{0.0326} & 0.6490 \\
& MedDiff-FM  & \textbf{0.8480} & \textbf{0.0130} & \textbf{0.0749} & 0.0648 & \textbf{0.7076} \\
\midrule
\multirow{3}{*}{Abdomen}
& DDPM        & 0.6009 & 0.0242 & 0.2792 & 0.0489 & 0.6998 \\
& MAISI       & 0.6905 & \textbf{0.0134} & \textbf{0.1373} & 0.0304 & 0.8169 \\
& MedDiff-FM  & \textbf{0.7439} & 0.0136 & 0.2595 & \textbf{0.0303} & \textbf{0.8524} \\
\midrule
\multirow{3}{*}{Overall}
& DDPM        & 0.7011 & 0.0201 & 0.1096 & 0.0740 & 0.6466 \\
& MAISI       & 0.6977 & 0.0175 & 0.0764 & \textbf{0.0348} & 0.7608 \\
& MedDiff-FM  & \textbf{0.7984} & \textbf{0.0125} & \textbf{0.0655} & 0.0409 & \textbf{0.8183} \\
\bottomrule
\end{tabular}
\label{tab:imagesynthesis}
\end{table*}

\subsubsection{Evaluation Metrics}
We utilize metrics widely used to assess synthesis quality and diversity, including the Multi-Scale Structural Similarity Index Measure (MS-SSIM)~\cite{wang2003multiscale}, Learned Perceptual Image Patch Similarity (LPIPS)~\cite{zhang2018unreasonable}, Fréchet Inception Distance (FID)~\cite{heusel2017gans}, and Maximum Mean Discrepancy (MMD)~\cite{gretton2012kernel}.
The feature extractor is a 3D pre-trained ResNet~\cite{he2016deep} from MedicalNet~\cite{chen2019med3d}.

To further measure the consistency between the generated images and the given anatomical structure conditions for image synthesis tasks. We extract the segmented anatomical structures of the generated images using the TotalSegmentator~\cite{wasserthal2023totalsegmentator}, and calculate the Dice coefficient (Dice) between these structures and the given anatomical conditions for the major organs. For the head and neck region, this includes the brain and skull; for the chest region, the lung, heart, and aorta; and for the abdomen region, the spleen, kidney, gallbladder, liver, stomach, pancreas, small bowel, duodenum, and colon.

To evaluate the distribution similarity between the generated lesion images and the real lesion images for the lesion generation and lesion inpainting tasks, we train segmentation models on real lesion images with nnU-Net~\cite{isensee2021nnu}. Then we utilize the trained segmentation models to segment the generated lesion images, and calculate the Dice between the segmentation results and ground-truths. We compare the Dice coefficients for real lesion images and generated lesion images to assess their distribution similarity.

For the image denoising task, we use the structural similarity index measure (SSIM)~\cite{wang2004image} and the peak signal-to-noise ratio (PSNR) to validate the image denoising performance, following Dn-Dp~\cite{liu2025diffusion}. For super-resolution, we also utilize SSIM and PSNR as evaluation metrics following TVSRN~\cite{yu2022rplhr}.

For anomaly detection, we adopt both image-level and pixel-level Area Under the Receiver Operating Characteristic Curve (AUROC), together with the per-region overlap (PRO) as demonstrated in~\cite{bao2024bmad}, as well as the Dice score with threshold 0.5 as a reference metric to assess the anomaly detection performance.

\subsection{Effectiveness Evaluation of Model Components}
To demonstrate the effectiveness of each model component, we utilize MedDiff-FM to accomplish the conditional synthesis of CT volumes with flexible sizes across the HaN, chest, and abdomen regions. Starting from the patch-level DDPM baseline, we incrementally introduce each component to evaluate its effectiveness.

The ablation study results are summarized in Table~\ref{tab:ablation}. From the perspective of generative metrics, the overall results demonstrate that incorporating multi-level integration significantly improves the performance of flexible-size image synthesis. The combination of position and position embedding further enhances image synthesis quality. The results across different anatomical regions show that, although the gains from combining position and position embedding are sometimes marginal, the overall improvements are effective. From the perspective of the Dice score in the last column, the higher Dice score indicates that the synthetic images are more consistent with the distribution of real CT images, which confirms the importance of each model component. In summary, these quantitative results highlight the effectiveness of each component in MedDiff-FM.

\begin{figure}[htbp]
\centering
\includegraphics[page=5, trim={1.5cm, 0cm, 1.5cm, 1.5cm}, clip, width=1.0\linewidth]{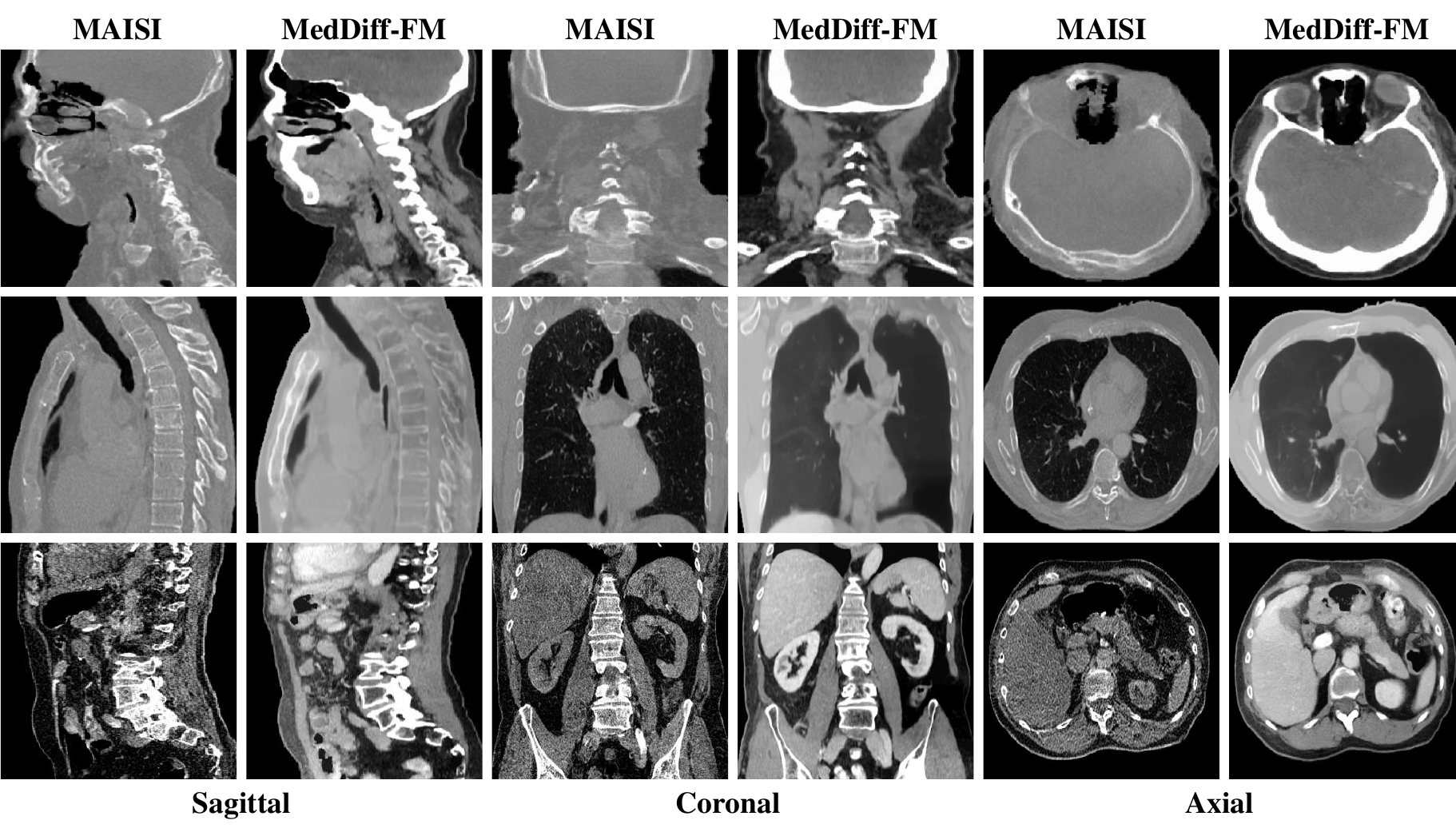}
\caption{Comparison visualization results of whole CT volume synthesis, including HaN, chest, and abdomen regions.}
\label{fig:imagesynthesis}
\end{figure}

\subsection{Evaluation of Downstream Tasks without Fine-tuning}
To validate the effectiveness of directly leveraging the pre-trained model in addressing downstream tasks, we evaluate MedDiff-FM on the image synthesis, image denoising, and anomaly detection tasks.

\subsubsection{Image Synthesis}
We compare MedDiff-FM with state-of-the-art methods on conditional CT volume synthesis of flexible sizes given anatomical regions and structures, including HaN, chest, and abdomen. We retrain DDPM with anatomical region and structure conditions on our pre-training dataset. We also compare with the pre-trained MAISI~\cite{guo2025maisi} model conditioned on anatomical structure segmentation masks, adapting our conditions to match the body region and anatomical structure formats of the MAISI model.

The quantitative comparison results are presented in Table~\ref{tab:imagesynthesis}. MedDiff-FM demonstrates overall superior performance in conditional CT image synthesis with flexible sizes. Although MAISI occasionally underperforms DDPM in terms of MS-SSIM, it consistently outperforms DDPM on other metrics, indicating its advantage in generating images with better feature representations. Although MedDiff-FM shows slightly inferior performance to MAISI in terms of abdominal FID and chest MMD, it achieves superior results for multi-level flexible-size image synthesis. The Dice scores in the last column are more reliable, as higher values indicate better alignment between the synthetic images and both the given conditions and the distribution of real CT images. The qualitative results of MAISI and MedDiff-FM are shown in Fig.~\ref{fig:imagesynthesis}, including the axial, coronal, and sagittal views. The CT volumes generated by MedDiff-FM exhibit better continuity and finer details. These results demonstrate that the multi-level integration and position embedding in MedDiff-FM successfully facilitate flexible-size image synthesis and the construction of multi-level spatial relationships.

\begin{table}[htbp]
\centering
\footnotesize
\caption{Image denoising performance on the Mayo 2016 dataset. We report the PSNR and SSIM improvements of different methods over their LDCT baselines.}
\begin{tabular}{clcc}
\toprule
& Method & PSNR $\uparrow$ & SSIM $\uparrow$ \\
\midrule
\multirow{2}{*}{\makecell{Supervised\\Method}}
& RED-CNN~\cite{chen2017low} & +4.55 & +0.051 \\
& CTformer~\cite{wang2023ctformer} & \textbf{+4.66} & +0.052 \\
\midrule
\multirow{5}{*}{\makecell{Unsupervised\\Method}}
& NLM~\cite{buades2005non} & +2.44 & +0.032 \\
& BM3D~\cite{dabov2007image} & +2.73 & +0.033 \\
& Noise2Sim~\cite{niu2022noise} & +4.22 & +0.050 \\
& Dn-Dp~\cite{liu2025diffusion} & +4.40 & +0.049 \\
& MedDiff-FM & +4.19 & \textbf{+0.059} \\
\bottomrule
\end{tabular}
\label{tab:imagedenoise}
\end{table}

\subsubsection{Image Denoising}
\begin{figure}[htbp]
\centering
\includegraphics[page=8, trim={9cm, 3cm, 12cm, 2.5cm}, clip, width=0.7\linewidth]{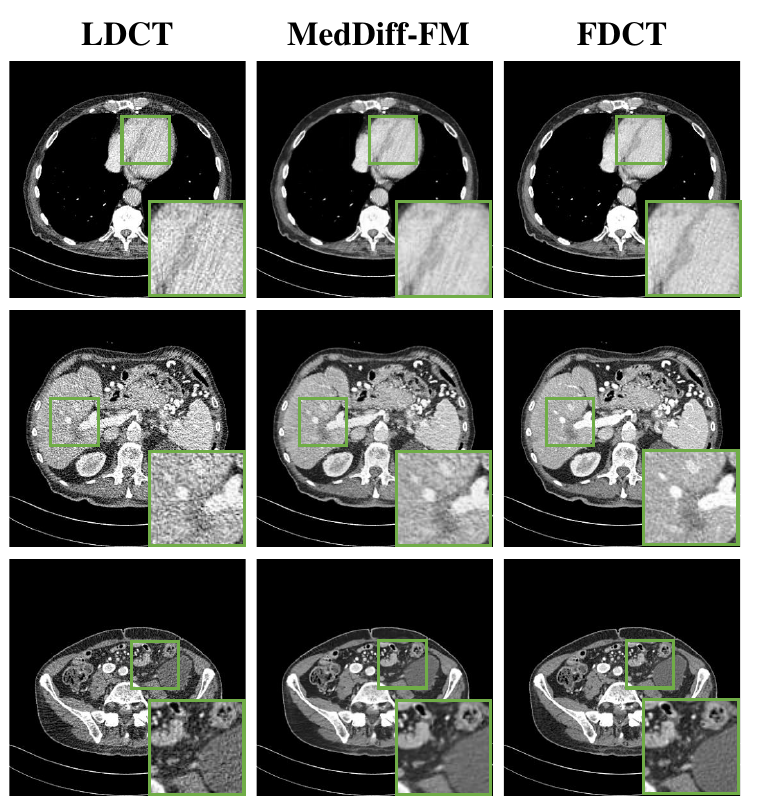}
\caption{The visualization results of image denoising task on the Mayo 2016 dataset. The window for displaying is [-160, 240] HU.}
\label{fig:imagedenoise}
\end{figure}

We evaluate the image denoising capabilities of MedDiff-FM on the Mayo 2016 dataset using the official 25\% dose CT images. We utilize the off-the-shelf MedDiff-FM for image denoising as introduced in Section~\ref{sec:imagedenoise}. The standard deviation of the additive noise $\sigma_n$ is estimated from the non-test cases and is approximately 0.15. And the regularization parameter $\lambda$ is set to 10.
The results of the comparison methods in Table~\ref{tab:imagedenoise} are cited from~\cite{liu2025diffusion}. Due to the different preferences for window width and level settings across methods, we report the PSNR and SSIM improvements of each method over its corresponding LDCT baseline under its respective window width and level, using the identical test cases. Specifically, the LDCT baseline under the window of $[-1024,3072]$ HU yields a PSNR of 40.41\,dB and an SSIM of 0.922, and the LDCT baseline under the window of $[-160,240]$ HU achieves a PSNR of 23.10\,dB and an SSIM of 0.757.
MedDiff-FM performs image denoising under the window of $[-160,240]$ HU. As shown in Table~\ref{tab:imagedenoise}, MedDiff-FM even significantly outperforms supervised comparison methods in terms of SSIM, while maintaining competitive PSNR performance. Unlike all other comparison methods, MedDiff-FM is applied to image denoising without additional training on the Mayo 2016 dataset, thus reducing the consumption of spatio-temporal resources. The qualitative results in Fig.~\ref{fig:imagedenoise} intuitively demonstrate that MedDiff-FM achieves effective denoising and enhances image quality, highlighting its strong capabilities in image denoising.

\begin{table}[htbp]
\centering
\footnotesize
\caption{Anomaly detection performance on the liver CT anomaly detection from the BMAD benchmark.}
\begin{tabular}{lcccc}
\toprule
Method & \makecell{Image\\AUROC} $\uparrow$ & \makecell{Pixel\\AUROC} $\uparrow$ & PRO $\uparrow$ & Dice $\uparrow$ \\ 
\midrule
DRAEM~\cite{zavrtanik2021draem} & 0.6995 & 0.8745 & 0.7929 & 0.0938 \\
MKD~\cite{salehi2021multiresolution} & 0.6072 & 0.9606 & 0.9108 & 0.1492 \\
RD4AD~\cite{deng2022anomaly} & 0.6038 & 0.9601 & 0.9029 & 0.1072 \\
STFPM~\cite{yamada2021reconstruction} & 0.6175 & 0.9118 & 0.9062 & 0.0887 \\
PaDiM~\cite{defard2021padim} & 0.5078 & 0.9094 & 0.7679 & 0.0450 \\
PatchCore~\cite{roth2022towards} & 0.6028 & 0.9643 & 0.8775 & 0.1049 \\
CFA~\cite{lee2022cfa} & 0.6200 & 0.9724 & 0.9275 & 0.1493 \\
CFLOW~\cite{gudovskiy2022cflow} & 0.5080 & 0.9241 & 0.8311 & 0.0758 \\
SimpleNet~\cite{liu2023simplenet} & 0.7228 & 0.9751 & 0.9107 & 0.1226 \\
MedDiff-FM & \textbf{0.8525} & \textbf{0.9760} & \textbf{0.9663} & \textbf{0.2744} \\
\bottomrule
\end{tabular}
\label{tab:anomalydetect}
\end{table}

\subsubsection{Anomaly Detection}
\begin{figure}[htbp]
\centering
\includegraphics[page=9, trim={8cm, 3cm, 13cm, 2.5cm}, clip, width=0.6\linewidth]{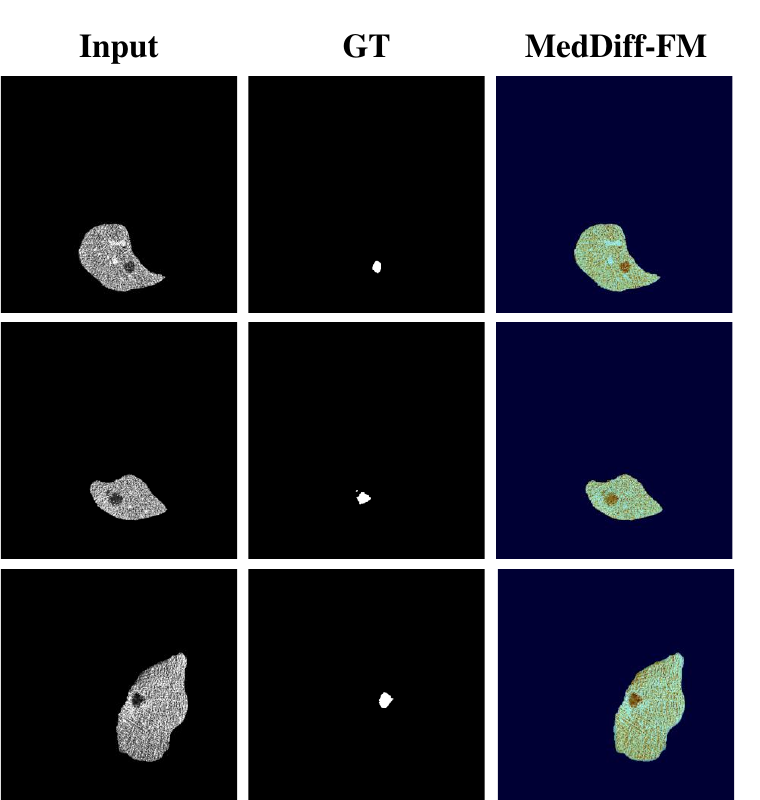}
\caption{Visualization of anomaly maps on the liver CT anomaly detection from the BMAD benchmark, where redness indicates higher anomaly scores.}
\label{fig:anomalydetect}
\end{figure}

We utilize pre-trained MedDiff-FM to accomplish the anomaly detection task on the liver CT dataset from the BMAD~\cite{bao2024bmad} benchmark. The test images consist of 1,493 2D slices, which we stack along the z-axis to form pseudo-3D volumes as inputs to MedDiff-FM during inference. As introduced in Section~\ref{sec:anomalydetect}, MedDiff-FM adds noise to abnormal CT scans at a fixed time step to disrupt tumor structures. Because small time steps struggle to corrupt tumor regions~\cite{wyatt2022anoddpm}, we diffuse the abnormal CT scans using a fixed time step of 950.
The quantitative results of the baseline methods in Table~\ref{tab:anomalydetect} are taken from BMAD~\cite{bao2024bmad}. In accordance with~\cite{bao2024bmad}, threshold-independent metrics are adopted for evaluation, with the Dice score being additionally used as a threshold-dependent reference metric. The quantitative results are summarized in Table~\ref{tab:anomalydetect}. The pre-trained MedDiff-FM outperforms state-of-the-art anomaly detection methods by a large margin, especially in terms of Image AUROC, PRO, and Dice. Fig.~\ref{fig:anomalydetect} visualizes the anomaly maps, where MedDiff-FM accurately localizes liver tumors. These results highlight the strong potential of MedDiff-FM for anomaly detection in medical imaging.

\subsection{Evaluation of Downstream Tasks with Fine-tuning}
To further demonstrate the capabilities of MedDiff-FM, we fine-tune it for super-resolution, lesion generation, and lesion inpainting tasks.

\begin{table}[htbp]
\centering
\footnotesize
\caption{Super-resolution performance on the RPLHR-CT dataset.}
\begin{tabular}{lcc}
\toprule
Method & PSNR $\uparrow$ & SSIM $\uparrow$ \\ 
\midrule
Bicubic & 33.508 & 0.902 \\
ResVox~\cite{ge2019stereo} & 37.946 & 0.932 \\
MPU-Net~\cite{liu2020multi} & 37.140 & 0.924 \\
SAINT~\cite{peng2020saint} & 38.019 & 0.933 \\
DA-VSR~\cite{peng2021vsr} & 36.672 & 0.924 \\
TVSRN~\cite{yu2022rplhr} & 38.609 & 0.936 \\
MedDiff-FM & 37.654 & 0.929 \\
\bottomrule
\end{tabular}
\label{tab:superres}
\end{table}

\begin{table*}[!t]
\centering
\footnotesize
\caption{Quantitative comparison of lesion generation performance across four datasets with different kinds of lesions.}
\begin{tabular}{clcccc>{\columncolor{lightgray}}c}
\toprule
Dataset & Method & MS-SSIM $\uparrow$ & LPIPS $\downarrow$ & FID $\downarrow$ & MMD $\downarrow$ & Dice $\uparrow$ \\
\midrule
\multirow{4}{*}{MSD-Lung}
& \textcolor{gray!90}{Real} & \textcolor{gray!90}{-} & \textcolor{gray!90}{-} & \textcolor{gray!90}{-} & \textcolor{gray!90}{-} & \textcolor{gray!90}{0.74} \\
& Med-DDPM & 0.4932 & 0.0473 & 0.2808 & 0.0592 & 0.24 \\
& MedDiff-FM (from scratch) & 0.6606 & 0.0135 & 0.2059 & 0.1018 & 0.16 \\
& MedDiff-FM (fine-tune) & 0.6731 & 0.0136 & 0.2332 & 0.0998 & 0.28 \\
\midrule
\multirow{4}{*}{MSD-Liver}
& \textcolor{gray!90}{Real} & \textcolor{gray!90}{-} & \textcolor{gray!90}{-} & \textcolor{gray!90}{-} & \textcolor{gray!90}{-} & \textcolor{gray!90}{0.71} \\
& Med-DDPM & 0.3030 & 0.0540 & 0.4795 & 0.1202 & 0.25 \\
& MedDiff-FM (from scratch) & 0.5991 & 0.0151 & 0.1304 & 0.0679 & 0.40 \\
& MedDiff-FM (fine-tune) & 0.6121 & 0.0138 & 0.1690 & 0.0727 & 0.48 \\
\midrule
\multirow{4}{*}{MED-LN}
& \textcolor{gray!90}{Real} & \textcolor{gray!90}{-} & \textcolor{gray!90}{-} & \textcolor{gray!90}{-} & \textcolor{gray!90}{-} & \textcolor{gray!90}{0.28} \\
& Med-DDPM & 0.7002 & 0.0170 & 0.1023 & 0.0508 & 0.01 \\
& MedDiff-FM (from scratch) & 0.7946 & 0.0059 & 0.3057 & 0.0538 & 0.01 \\
& MedDiff-FM (fine-tune) & 0.7982 & 0.0074 & 0.1195 & 0.0183 & 0.22 \\
\midrule
\multirow{4}{*}{ABD-LN}
& \textcolor{gray!90}{Real} & \textcolor{gray!90}{-} & \textcolor{gray!90}{-} & \textcolor{gray!90}{-} & \textcolor{gray!90}{-} & \textcolor{gray!90}{0.51} \\
& Med-DDPM & 0.4430 & 0.0336 & 0.2574 & 0.0708 & 0.39 \\
& MedDiff-FM (from scratch) & 0.5267 & 0.0304 & 0.1765 & 0.0570 & 0.32 \\
& MedDiff-FM (fine-tune) & 0.5802 & 0.0211 & 0.2425 & 0.0594 & 0.50 \\
\bottomrule
\end{tabular}
\label{tab:lesiongen}
\end{table*}

\subsubsection{Super-resolution}
MedDiff-FM is fine-tuned for volumetric super-resolution using the training set from the RPLHR-CT dataset~\cite{yu2022rplhr}. The fine-tuned MedDiff-FM is then evaluated on the test set of 100 cases from the same dataset. As demonstrated in Section~\ref{sec:superres}, the standard deviation of the additive noise $\sigma_n$ is set to 1.0 as discussed in~\cite{yu2022rplhr} and the regularization parameter $\lambda$ is set to 1. The $t_{start}$ starts from $T$, with 100 Neural Function Evaluations (NFEs) using uniform skipping. The scale factor $sf$ is 5 in depth dimension since the real-paired thick-CTs have a slice thickness of 5\,mm, while the thin-CTs have a slice thickness of 1\,mm.

\begin{figure}[htbp]
\centering
\includegraphics[page=10, trim={7cm, 4cm, 10cm, 3cm}, clip, width=0.7\linewidth]{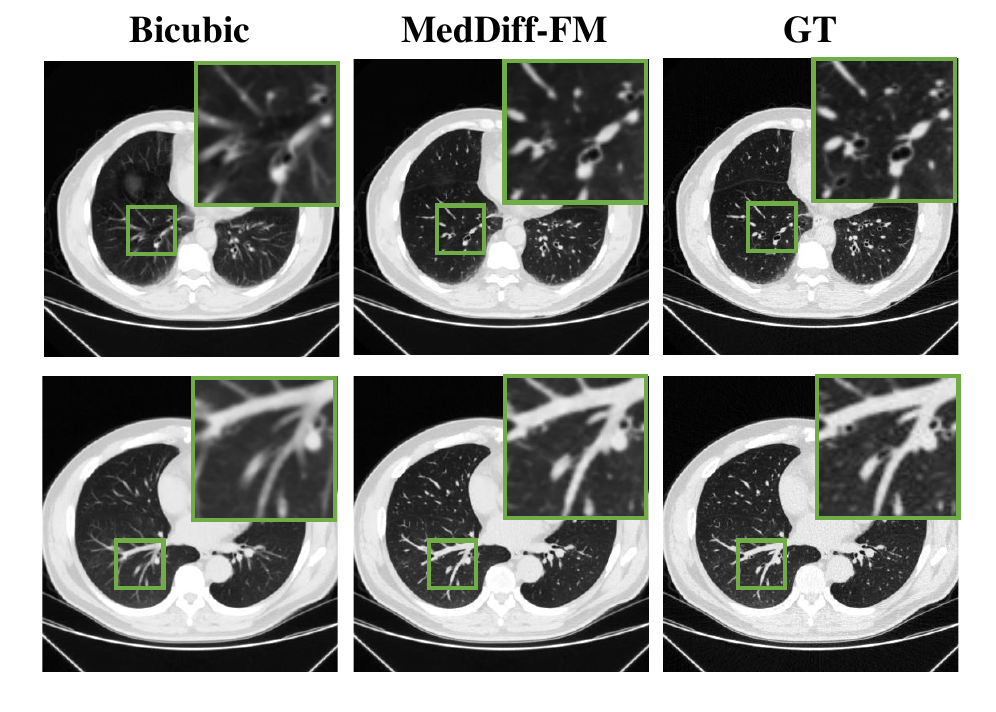}
\caption{Visualization of super-resolution results on the real-paired RPLHR-CT dataset. The window for displaying is [-1024, 150] HU.}
\label{fig:superres}
\end{figure}

The results of the comparison methods in Table~\ref{tab:superres} are reported as in~\cite{yu2022rplhr}. As shown in Table~\ref{tab:superres}, MedDiff-FM achieves competitive performance on the real-paired RPLHR-CT dataset for the volumetric super-resolution task. We take advantage of DiffPIR~\cite{zhu2023denoising} to adapt MedDiff-FM to this inverse problem. DiffPIR requires a known degradation operator, i.e., a known downsampling operator. However, for real-paired thick-CTs and thin-CTs, we can only approximate the downsampling operator, which differs from the actual degradation process. As a result, the performance of MedDiff-FM is somewhat limited. Nevertheless, as visualization in Fig.~\ref{fig:superres}, MedDiff-FM achieves promising super-resolution results that are close to ground-truth images.

\subsubsection{Lesion Generation}
We fine-tune MedDiff-FM for lesion generation using four datasets containing various types of lesions: MSD-Lung, MSD-Liver, MED-LN, and ABD-LN. We compare the fine-tuned MedDiff-FM with Med-DDPM~\cite{dorjsembe2024conditional} and MedDiff-FM trained from scratch. For fair comparison, we train Med-DDPM conditioned on anatomical structure masks and lesion masks.
The quantitative results on four lesion datasets are listed in Table~\ref{tab:lesiongen}. Taking into account the results across all generative metrics, the MedDiff-FM framework exhibits notable advantages over the Med-DDPM method, although the fine-tuned MedDiff-FM does not consistently achieve the best performance in LPIPS, FID, and MMD. It is important to note that these generative metrics assess the quality of synthetic lesion images from a holistic perspective. However, for lesion generation, it is more crucial to emphasize the quality of the lesions themselves. To provide a better assessment of synthetic lesion quality, we employ segmentation models trained on real lesion data to segment synthetic lesions and compute the Dice score. The Dice score results in the last column demonstrate that the fine-tuned MedDiff-FM significantly surpasses other methods in high-quality lesion generation.

\begin{figure}[htbp]
\centering
\includegraphics[page=6, trim={3.5cm, 0cm, 2cm, 0cm}, clip, width=0.95\linewidth]{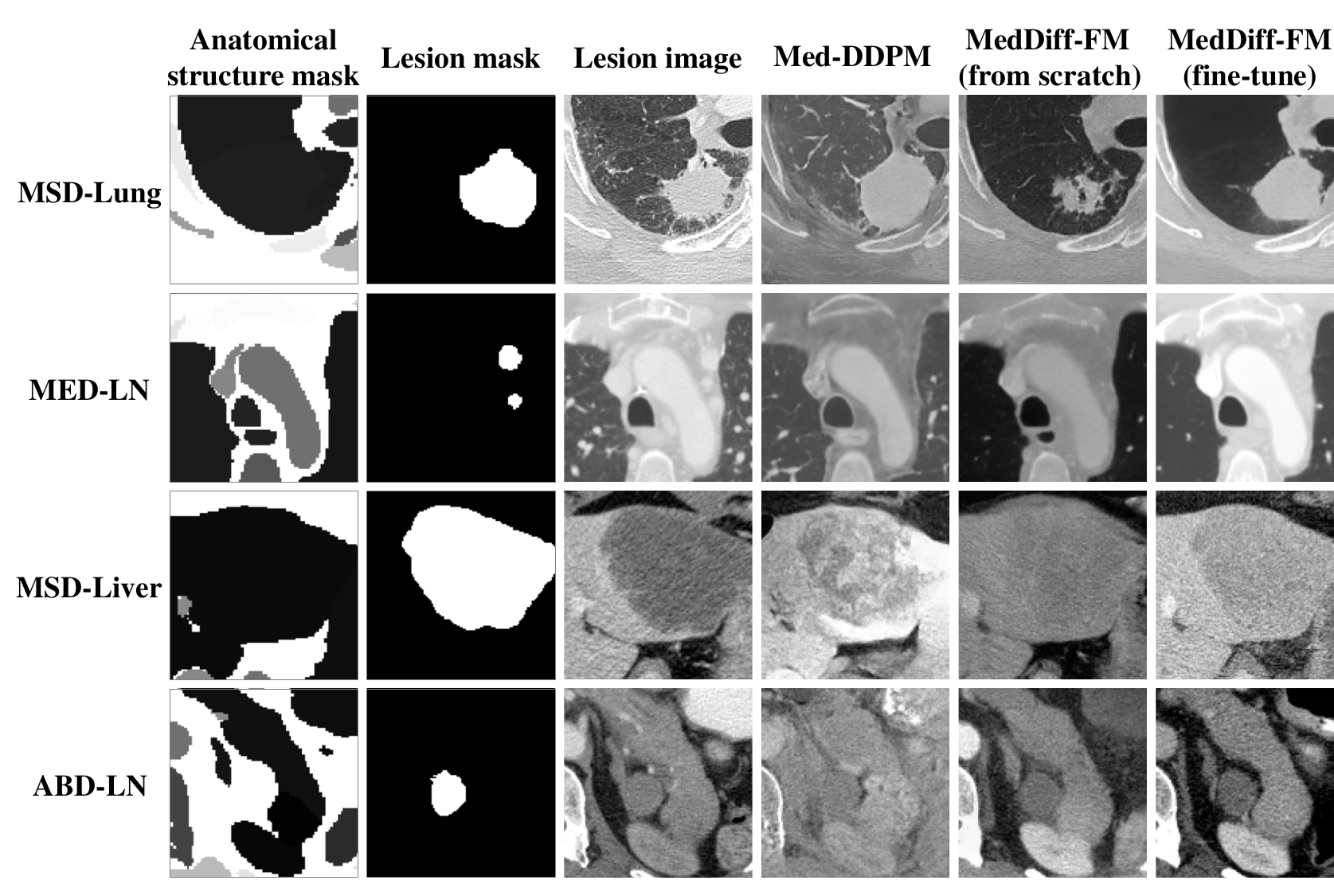}
\caption{Examples of lesion generation results given specific anatomical structure masks and lesion masks. The lesion images are provided for reference only.}
\label{fig:lesiongen}
\end{figure}

The visualization of lesion generation results in Fig.~\ref{fig:lesiongen} suggests that MedDiff-FM achieves superior performance in both image synthesis and lesion generation. Furthermore, qualitative visualization and quantitative Dice scores on the MED-LN dataset indicate that generating mediastinal lymph nodes is particularly challenging. When trained on the limited MED-LN data, the comparison models exhibit poor performance. In contrast, fine-tuning the pre-trained MedDiff-FM on the same limited data substantially improves the quality of mediastinal lymph node generation.

\begin{table}[!t]
\centering
\footnotesize
\caption{Quantitative comparison of lesion inpainting performance.}
\begin{tabular}{clc}
\toprule
Dataset & Method & Dice $\uparrow$ \\
\midrule
\multirow{4}{*}{MSD-Lung}
& \textcolor{gray!90}{Real} & \textcolor{gray!90}{0.74} \\
& DiffTumor & 0.44 \\
& MedDiff-FM (from scratch) & 0.42 \\
& MedDiff-FM (fine-tune) & 0.77 \\
\midrule
\multirow{4}{*}{MSD-Liver}
& \textcolor{gray!90}{Real} & \textcolor{gray!90}{0.71} \\
& DiffTumor & 0.67 \\
& MedDiff-FM (from scratch) & 0.71 \\
& MedDiff-FM (fine-tune) & 0.71 \\
\midrule
\multirow{4}{*}{MED-LN}
& \textcolor{gray!90}{Real} & \textcolor{gray!90}{0.28} \\
& DiffTumor & 0.01 \\
& MedDiff-FM (from scratch) & 0.30 \\
& MedDiff-FM (fine-tune) & 0.34 \\
\midrule
\multirow{4}{*}{ABD-LN}
& \textcolor{gray!90}{Real} & \textcolor{gray!90}{0.51} \\
& DiffTumor & 0.45 \\
& MedDiff-FM (from scratch) & 0.51 \\
& MedDiff-FM (fine-tune) & 0.53 \\
\bottomrule
\end{tabular}
\label{tab:lesioninpaint}
\end{table}

\subsubsection{Lesion Inpainting}
\begin{figure}[htbp]
\centering
\includegraphics[page=7, trim={5cm, 0cm, 5cm, 0cm}, clip, width=0.85\linewidth]{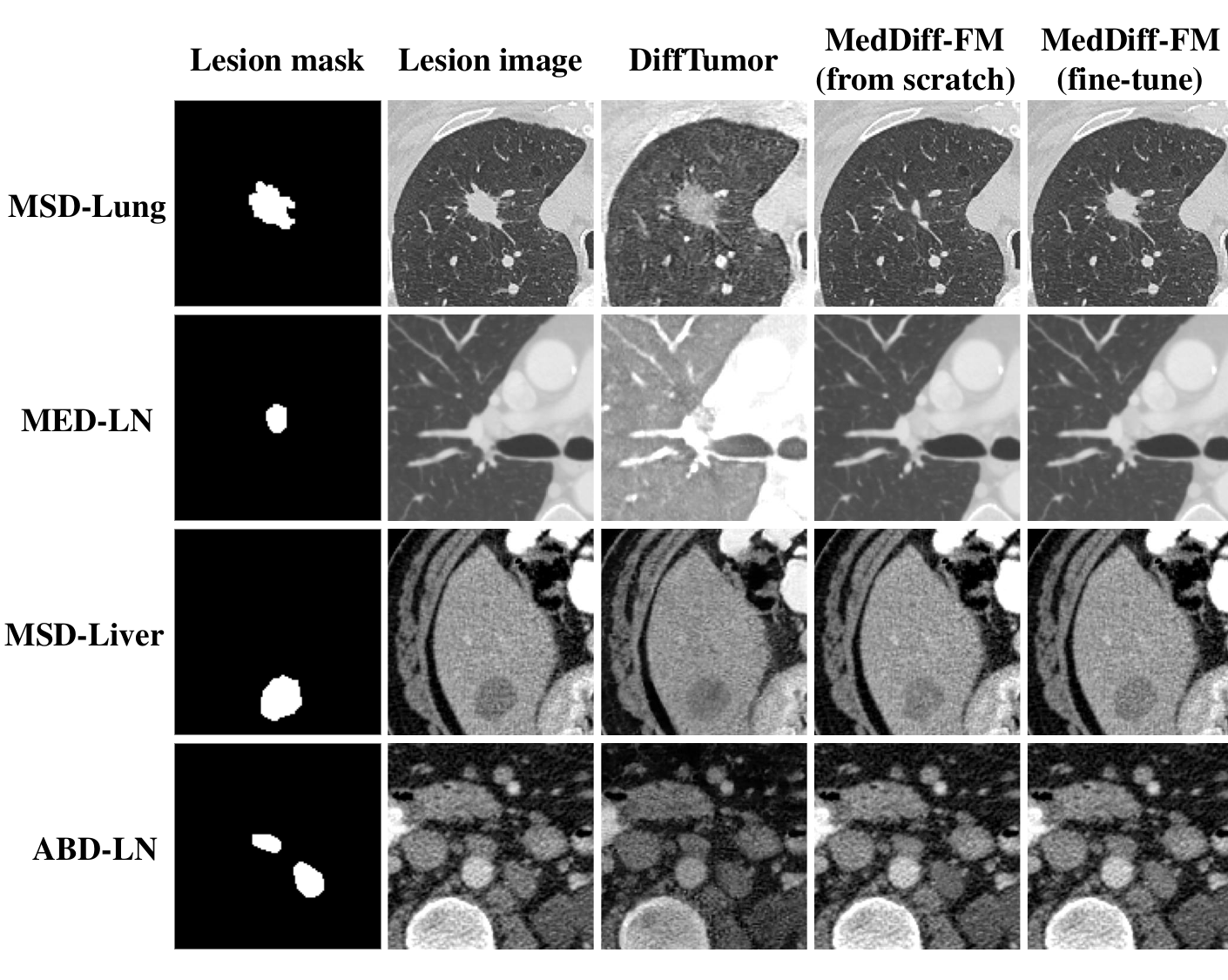}
\caption{Lesion inpainting samples conditioned on lesion masks, with lesion images shown for reference only.}
\label{fig:lesioninpaint}
\end{figure}

To further evaluate the realism and quality of the synthetic lesions themselves, we apply the fine-tuned MedDiff-FM to the lesion inpainting task. DiffTumor~\cite{chen2024towards} is designed for tumor synthesis on healthy abdominal organs. To ensure a fair comparison, we retrain the diffusion model of DiffTumor on each dataset while using its released autoencoder model pre-trained on large-scale abdominal data.
Since image inpainting does not change the holistic structure of the original image, we consider it unnecessary to evaluate the quality of the holistic image using general generative metrics. We employ segmentation models trained on real lesions to segment inpainted lesions and compute the Dice score. The quantitative results on four lesion datasets are presented in Table~\ref{tab:lesioninpaint}, and the qualitative results are illustrated in Fig.~\ref{fig:lesioninpaint}. Given that the autoencoder is pre-trained on abdominal data and the lesion synthesis on the MED-LN dataset is particularly challenging, DiffTumor fails to generate mediastinal lymph nodes. In summary, both qualitative and quantitative results indicate that the synthetic lesions of MedDiff-FM are more consistent with the distribution of real lesions, which validates the effectiveness of MedDiff-FM.

\section{Discussion}
The proposed MedDiff-FM covers multiple anatomical regions and handles various downstream tasks, demonstrating robust generalization capabilities. The architecture of the diffusion foundation model is flexible, supporting multi-level integrated image processing. However, there are some limitations in this work. First, we utilize anatomical structures to control local details in order to generate higher-quality medical images, which limits flexibility. Future work could explore more flexible conditioning mechanisms, such as textual prompts, which enable interactive image synthesis. Second, while MedDiff-FM is specifically designed for CT images, it highlights the potential for diffusion-based foundation models to be extended to other medical imaging modalities, such as MRI and PET. Furthermore, for the whole CT volume generation, the patch-based sliding window inference strategy, in conjunction with the progressive denoising process, imposes substantial computational burdens. To address this, future work could explore methods like consistency models~\cite{song2023consistency} to accelerate the sampling process while maintaining a balance between inference efficiency and image quality.

\section{Conclusion}
In conclusion, this paper presents MedDiff-FM, a diffusion-based foundation model that deals with a wide range of medical image tasks. MedDiff-FM utilizes 3D CT images from diverse publicly available datasets and focuses on multiple anatomical regions, overcoming the limitations of previous work that were constrained by specific anatomical regions and particular tasks. MedDiff-FM is capable of handling multi-level integrated image processing with position embedding to build multi-level spatial relationships, and using anatomical structures and regions as conditions. The pre-trained diffusion foundation model can seamlessly perform tasks such as image denoising, anomaly detection, and image synthesis. Furthermore, MedDiff-FM deals with super-resolution, lesion generation, and lesion inpainting through efficient fine-tuning via ControlNet with task-specific conditions. The experimental results highlight the effectiveness of MedDiff-FM, making it a valuable tool for various medical image applications.

\bibliographystyle{IEEEtran}
\bibliography{ref}

\end{document}